\documentclass[10pt,twocolumn,letterpaper]{article}

\usepackage{wacv}
\usepackage{times}
\usepackage{epsfig}
\usepackage{graphicx}
\usepackage{subfigure}
\usepackage{fancyhdr}
\usepackage{color}
\usepackage{amsmath,amssymb,amsfonts,amscd}
\usepackage{cite}

\usepackage[pagebackref=true,breaklinks=true,letterpaper=true,colorlinks,bookmarks=false]{hyperref}

\wacvfinalcopy 

\def\wacvPaperID{365} 

\ifwacvfinal\fi
\begin{document}

\title{Structured Hough Voting for Vision-based Highway Border Detection}

\author{Zhiding Yu$^{\star}$, Wende Zhang$^{\dagger}$, B. V. K. Vijaya Kumar$^{\star}$, Dan Levi$^{\ddagger}$\\
$^{\star}$Department of Electrical and Computer Engineering, Carnegie Mellon University\\
$^{\dagger}$Electrical and Controls Systems Research Lab, General Motors Company\\
$^{\ddagger}$Advanced Technical Center - Israel, General Motors Company\\
\tt\small {yzhiding@andrew.cmu.edu, \{wende.zhang,dan.levi\}@gm.com, kumar@ece.cmu.edu}
}

\maketitle
\ifwacvfinal\thispagestyle{empty}\fi

\begin{abstract}
We propose a vision-based highway border detection algorithm using structured Hough voting. Our approach takes advantage of the geometric relationship between highway road borders and highway lane markings. It uses a strategy where a number of trained road border and lane marking detectors are triggered, followed by Hough voting to generate corresponding detection of the border and lane marking. Since the initially triggered detectors usually result in large number of positives, conventional frame-wise Hough voting is not able to always generate robust border and lane marking results. Therefore, we formulate this problem as a joint detection-and-tracking problem under the structured Hough voting model, where tracking refers to exploiting inter-frame structural information to stabilize the detection results. Both qualitative and quantitative evaluations show the superiority of the proposed structured Hough voting model over a number of baseline methods.
\end{abstract}

\section{Introduction}
Detecting road borders has broad applications in future autonomous vehicles and intelligent transportation systems as an important component of scene understanding. It can provides cues about road structure that benefit motion planning and cruise behavior control. In autonomous driving, the detection of road border is often done by GPS, high quality road map and some other active sensors such as Radar and Lidar. This, however, can sometimes be limited by the accuracy of GPS positioning signal as well as the resolution of active sensors. A natural question is whether we can address the problem with computer vision.

Besides detecting the physical road border, part of our task also includes robustly detecting the shoulder region in order to provide necessary maneuver guidance for future autonomous driving systems. In the United States, highways often contain a so-called ``shoulder region'' usually defined as the region between the outer-most solid lane marking and the physical road border. This shoulder region serves as a buffer zone before the physical limit of the road. Non-emergency vehicles are mostly not allowed to drive on shoulder regions. But under emergency conditions, they may be allowed on shoulder regions for purposes such as evasive maneuver and emergency parking.

\begin{figure}[t]
  \centering
  \label{Fig1a}
   \includegraphics[height=3cm]{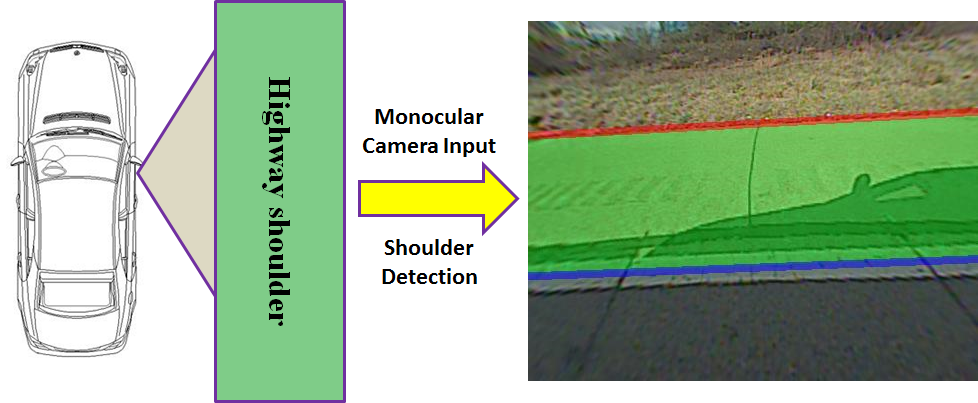}
\caption{Examples of the proposed system. The red line indicates the physical road border while the blue line is the lane marking. They together defined the green shoulder region.}\label{Fig1}
\end{figure}

We start from the most basic setting where a side-view production camera\footnote{Here production cameras refers to the cameras that have been mass produced in the transportation industry, often with low cost but also with relatively low image resolution and quality.} looks out from the right side of a vehicle, as illustrated in Fig.\ref{Fig1}. The detection problem is approached as the problem of detecting the road border (e.g., guard rails, concrete barriers, etc.) and detecting the closest lane marking on the shoulder-side of the vehicle. The expectation is that the algorithm will return an estimate of the drivable regions before the physical road border. If the vehicle is on the right-most lane, the shoulder region is expected to be detected, while if the vehicle is on inner lanes, the detection also include available lanes on the right. We will assume that the shoulder is on the right side of the car, but we can easily apply our approach to the shoulder on the left side provided that the side camera is mounted on the left side of the vehicle. We emphasize that the current system is just one part of a future surround view system where side-view detection and front-view detection can jointly support and improve each other. The current system can also benefit future autonomous driving systems by providing information about road structure. Our goal is to achieve robust detection within 0.5 to 6 meters detection range, while being able to handle various challenging scenarios including strong shadows, diverse border types/appearances and complicated scenarios such as highway entrances and exits.

In the United States, typical highway borders can be classified into three types: concrete barrier, guard rail and soft shoulder. Given this observation, our key assumption is that the types of highway border are not as diverse as the borders in urban or other uncontrolled scenarios and can be learned from a set of labelled images. This assumption, however, by no means makes the problem trivial: the border and shoulder detection problem still needs to address the limited resolution challenge, meet the required speed constraints, deal with complicated border situations and contend with a number of other challenges such as strong shadows, dynamic appearances and other patterns that look like borders.

We show that our problem can be theoretically formulated as a joint detection-and-tracking problem under a graphical model called ``structured Hough voting''. Our contribution in this paper lies in the fact that the proposed structured Hough voting model exploits a variety of inter-frame and intra-frame structural information to achieve very robust performance, while using multiple candidate hypotheses and mode selection to retain necessary flexibility. We will show that the proposed model performs very well on the highway border and shoulder detection problem.

\section{Related work}
The problem of vision-based scene understanding for autonomous driving has been widely studied. Many seek to address the problem of general object detection, such as the detection of pedestrians \cite{25}, bicycles \cite{44}, motorcycles and vehicles \cite{22}. There has also been a considerable amount of work regarding scene parsing where each pixel/superpixel in an image is labeled with a certain object class \cite{12,41}. However, most of them focus on the understanding of general objects and the algorithms are often not possible to run in real-time\footnote{by ``real time'' we mean at least 10 frames per second.} on a regular CPU.

Some relevant works try to understand the structure of road and these works are indispensable parts of the autonomous driving system. Considerable effort has been devoted to automatically detect roads \cite{2,6} and lane markings \cite{23}, or to find vanishing points \cite{20}. Others have addressed the problem by exploring more capable sensors, including stereo vision sensors \cite{30} and Lidar sensors \cite{38}. These sensors provide extra depth information which makes the tasks considerably easier. Thus they have been adopted in some autonomous systems \cite{33}. While these sensors are able to provide more information than monocular cameras, their costs are often very high.

The problem of road border detection has been previously addressed \cite{13,14,15,16,17,18} but none has addressed the highway scenario where border detection can become particularly difficult with concrete barriers due to their textureless nature. In addition, most of them focus on features and detectors while our work also presents a novel robust model.

\section{Proposed model}
Given a video from the side camera, we investigate both the inter-frame and intra-frame structural information instead of performing Hough voting independently for each frame. Our high-level intuition here is that these structural cues are the key to robust performance. To utilize such cues we formulate our model under a conditional random field (CRF). We shall see that independent Hough voting corresponds to unary prediction in our CRF model, returning how likely the hypotheses are. The inter-frame and intra-frame structural information corresponds to pairwise potentials, introducing additional constraints to refine the results.

\subsection{Hough voting background}
Geometrically, a straight line in a 2-D $(x,y)$ space can be represented by the following equation:
\begin{equation}
\sin(\theta)y+\cos(\theta)x-r = 0,
\end{equation}
where $r$ is the algebraic distance between the line and the origin, $\theta$ is the angle of the vector orthogonal to the line. Voting points are the points indicating where a line should be. Given a set of voting points whose coordinates are $\{(x_i,y_i)|i=1,...,N\}$, the voting weight is defined as:
\begin{equation}\label{weighted_hough}
v(\theta,r) = \sum_{i=1}^{N}w_{i}\exp(-\frac{(\sin(\theta)y_i+\cos(\theta)x_i-r)^2}{2\sigma^2}).
\end{equation}
where $w_i$ is the weight associated with each voting point. $\sigma$ is the bandwidth parameter that adjusts how sensitive voting is to relatively far away voting points and is empirically fixed as 5 in this work for its good performance.

Let $(\theta, r)\triangleq\mathbf{h}$, conventional Hough voting seeks to find a hypothesis $\mathbf{h}$ that maximizes the voting weight:
\begin{equation}
\mathbf{h^{*}} = \arg\max_{\mathbf{h}}v(\mathbf{h}).
\end{equation}
Finding $\mathbf{h^{*}}$ with continuous optimization is difficult as it is non-convex. A common way is to discretize $\mathbf{h}$ and search for the one with the maximum vote weight. The remaining questions are: 1. How are the voting points defined? 2. How to obtain these voting points?

\begin{figure}[t,b,h]
  \centering
  \subfigure{
    \includegraphics[height = 2.4cm]{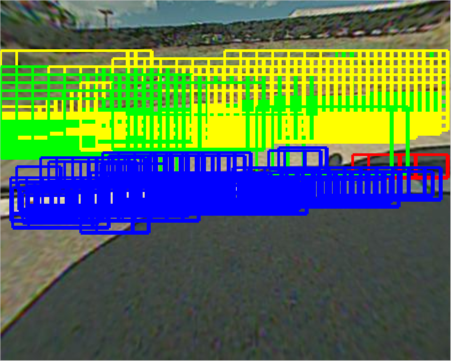}}
  \subfigure{
    \includegraphics[height = 2.4cm]{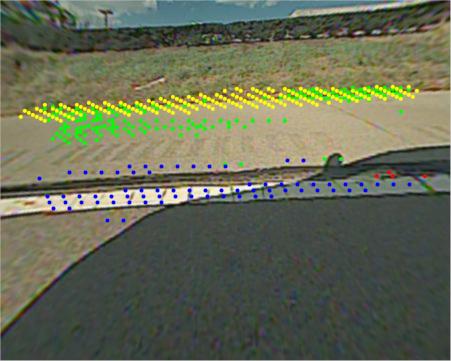}}
  \subfigure{
    \includegraphics[height = 2.4cm]{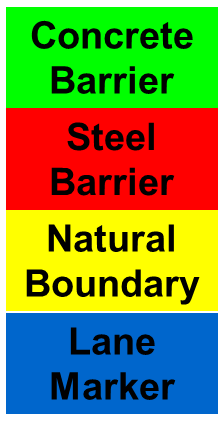}}
\caption{Examples of triggered detectors and their voting points.}\label{Fig5}
\end{figure}

\textbf{Type 1 voting points:} Voting points are returned by triggered border or lane marking detectors. Here we adopt scanning window detection where each detected positive window returns an equally weighted voting point estimating where the border/lane marking is\footnote{We will illustrate how to train these detectors later.}. We deliberately allow multiple dense triggers (See Fig. \ref{Fig5}) and the returned voting points are the main source of detection information.

\textbf{Type 2 voting points:} These are voting points whose coordinates are simply the coordinates of all pixels, while the Hough voting is now weighted by the gradient of each pixel. The intuition is obvious: borders and lane markings often have relatively strong vertical gradients. In case where detectors have failed and very few Type 1 voting points are returned, estimating with gradient might be a good approximate strategy to obtain a reasonable result.

\subsection{The structured Hough voting model}
We seek to treat the Hough voting hypotheses jointly, exploring their probabilistic and structured relations with a graphical model\footnote{\cite{48} also adopted the term ``structured Hough voting'' but is different.}. CRF is a highly suitable model for our problem. Suppose $\mathbf{X}$ denotes the aggregate of observations (i.e., the coordinates and weights of voting points) in all the frames and $\mathbf{H}$ the aggregate of all the Hough hypotheses, CRF discriminatively defines the joint posterior probability $P(\mathbf{H}|\mathbf{X})$ and the inference of CRF is to find the joint hypothesis configuration that maximizes $P(\mathbf{H}|\mathbf{X})$:
\begin{equation}\label{CRF_Infer}
\mathbf{H^{*}}=\mathop{\arg\max}_{\mathbf{H}}P(\mathbf{H}|\mathbf{X}),
\end{equation}

Different from many conventional models, our structured Hough voting model first generates three different candidate hypotheses\footnote{Here we mean both generating three border candidate hypotheses and three lane marking candidate hypotheses.} in every frame (except the initial one). We will show how they are generated later. The intuition is that we want to handle certain situations where: 1. Borders and lane markings can ``jump'' suddenly due to entrances and lane changes and 2. Failures of border/lane marking detectors can return very few Type 1 voting points. We hope that at least one of them can return a good ``guess'' of the border/lane marking in every situation and that the model can appropriately select the best expert.

Once the candidate hypotheses are obtained, the model selects the best one as the detection output. Finally, we also check whether the chosen hypotheses for border and lane marking in each frame have violated certain structure restrictions (e.g., whether they have intersected). If violations occur, perturbation is performed on the border hypothesis candidates to guarantee that such structural restrictions are followed. Again, one of the perturbated hypothesis candidates is selected as the final detection result.

Let ``bd'' and ``ln'' denote ``border'' and ``lane marking'' for short, we define the following notations to better describe our model\footnote{A hat symbol indicates the hypothesis is a candidate one.}. In the $i$th ($i\geq2$) frame, we have:\\
\textbf{Bd candidate hypotheses:} $\mathbf{\hat{H}}_{bd,i} \triangleq \{\mathbf{\hat{h}}_{bd1,i}, \mathbf{\hat{h}}_{bd2,i}, \mathbf{\hat{h}}_{bd3,i}\}$\\
\textbf{Ln candidate hypotheses:} $\mathbf{\hat{H}}_{ln,i} \triangleq \{\mathbf{\hat{h}}_{ln1,i}, \mathbf{\hat{h}}_{ln2,i}, \mathbf{\hat{h}}_{ln3,i}\}$\\
\textbf{Selected bd/ln hypotheses:} $\mathbf{h}_{bd,i} / \mathbf{h}_{ln,i}$\\
\textbf{Observations (bd, ln, grad):} $\mathbf{x}_{i}\triangleq \{\mathbf{x}_{bd,i}, \mathbf{x}_{ln,i}, \mathbf{x}_{grad,i}\}$.\\

\begin{figure}[t,b,h]
  \centering
  \includegraphics[height = 3.9cm]{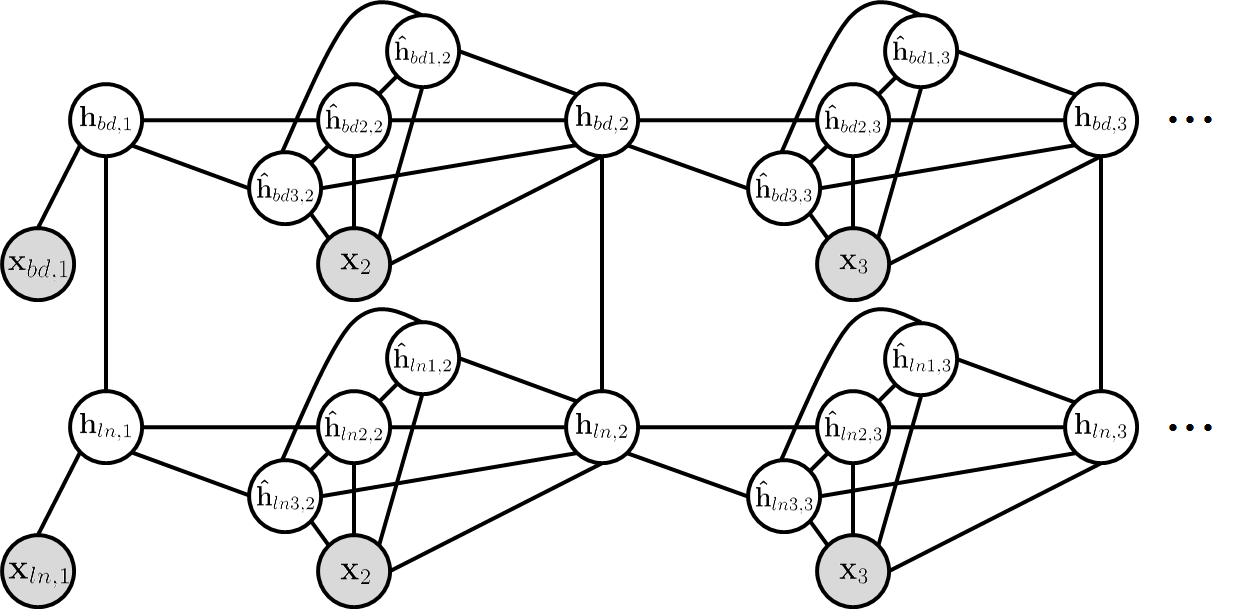}
\caption{The graphical model of the structured Hough voting.}\label{Model}
\end{figure}

Let $N\geq 2$ be the number of video frames. We model the log CRF conditional probability as a set of potentials\footnote{Here the potentials are functions with the bd/ln (candidate/output) hypotheses being the variables. They are modeled in such a way that a larger function value generally indicates better hypothesis configurations.}:
\small{
\begin{equation}
\begin{split}
&\log P(\mathbf{H}|\mathbf{X}) = \overbrace{\phi(\mathbf{h}_{bd,1},\mathbf{x}_{bd,1})}^{\text{Bd Hough voting}} + \overbrace{\phi(\mathbf{h}_{ln,1},\mathbf{x}_{ln,1})}^{\text{Ln Hough voting}}\\
&+\sum_{i=2}^{N}\overbrace{\Phi_{bd}(\mathbf{h}_{bd,i-1},\mathbf{\hat{H}}_{bd,i},\mathbf{x}_{i})}^{\text{Candidate bd hypo generation}} + \sum_{i=2}^{N}\overbrace{\Phi_{ln}(\mathbf{h}_{ln,i-1},\mathbf{\hat{H}}_{ln,i},\mathbf{x}_{i})}^{\text{Candidate ln hypo generation}}\\
&+\sum_{i=2}^{N}\overbrace{\Psi_{bd}(\mathbf{\hat{H}}_{bd,i},\mathbf{h}_{bd,i},\mathbf{x}_{i})}^{\text{Bd mode selection}} + \sum_{i=2}^{N}\overbrace{\Psi_{ln}(\mathbf{\hat{H}}_{ln,i},\mathbf{h}_{ln,i},\mathbf{x}_{i})}^{\text{Ln mode selection}}\\
&+\sum_{i=1}^{N}\overbrace{\Omega(\mathbf{h}_{bd,i},\mathbf{h}_{ln,i})}^{\text{Coupled structure}}\\
&-\log Z(\mathbf{X}).
\end{split}
\end{equation}
}

\begin{figure}[t,b,h]
  \centering
  \subfigure[]{
  \includegraphics[height = 2.4cm]{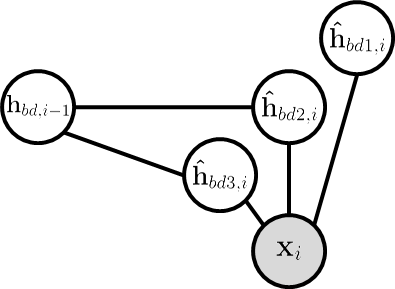}}\quad
  \subfigure[]{
  \includegraphics[height = 2.4cm]{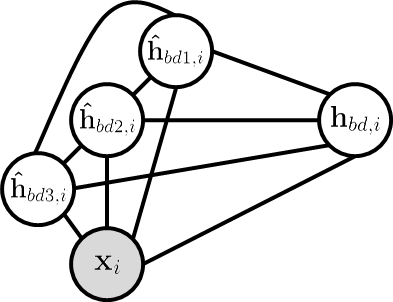}}\quad
  \subfigure[]{
  \includegraphics[height = 1.8cm]{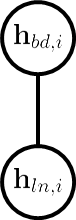}}
\caption{Some decomposed parts of the graphical: (a) Candidate hypothesis generation unit. (b). Mode selection potential. (c) Coupled structure potential.}\label{Model_Decompose}
\end{figure}

The graphical model is shown in Fig. \ref{Model}. Let $\Phi_{bd,i}$ denote $\Phi_{bd}(\mathbf{h}_{bd,i-1},\mathbf{\hat{H}}_{bd,i},\mathbf{x}_{i})$, and similarly for $\Phi_{ln,i}$, $\Psi_{bd,i}$, $\Psi_{ln,i}$ and $\Omega_{i}$, We will give detailed definition and intuition for each term in our model. To better illustrate each term, we decompose the model and show them in Fig. \ref{Model_Decompose}.

\subsubsection{Candidate hypotheses generation unit}
The term $\Phi_{bd}(\mathbf{h}_{bd,i-1},\mathbf{\hat{H}}_{bd,i},\mathbf{x}_{i})$ seeks to generate multiple candidate border hypotheses based on the observations in current frame and the selected hypotheses in the previous frame. The intuition here is that the first hypothesis candidate $\mathbf{\hat{h}}_{bd1,i}$ is generated by performing unconstrained\footnote{therefore there is no associated pairwise potential} Hough voting with Type 1 bd voting points. It is able to discover sudden border changes. The second candidate is also generated by Hough voting with Type 1 bd voting points, but is additionally constrained (smoothed) by the previous frame. The third candidate concerns the constrained Hough voting with Type 2 voting points (image gradients). It specifically handles the case of very few returned Type 1 voting points due to occlusions and faded lane markings. The graphical representation is shown in Fig. \ref{Model_Decompose} (a).

Note that the unit is not a clique potential. However, it is a composition of a set of potential functions:
\begin{equation}
\begin{split}
\Phi_{bd,i}& =\phi(\mathbf{\hat{h}}_{bd1,i},\mathbf{x}_{bd,i})\\
&+\phi(\mathbf{\hat{h}}_{bd2,i},\mathbf{x}_{bd,i})+ \varphi_{bd}(\mathbf{h}_{bd,i-1},\mathbf{\hat{h}}_{bd2,i})\\
&+\phi(\mathbf{\hat{h}}_{bd3,i},\mathbf{x}_{grad,i}) + \varphi_{bd}(\mathbf{h}_{bd,i-1},\mathbf{\hat{h}}_{bd3,i}),
\end{split}
\end{equation}
where $\phi$ is the Hough voting function defined in Eq. (\ref{weighted_hough}). Let $\Delta \theta \triangleq |\theta_{1}-\theta_{2}|$ and $\Delta r \triangleq |r_{1}-r_{2}|$, $\varphi$ is the inter-frame pairwise potential defined as a binary loss function:
\begin{equation}
\begin{split}
\varphi_{bd}&(\mathbf{h}_{1},\mathbf{h}_{2})=\\
&\left\{
\begin{aligned}
0, &\quad \text{if} \quad \Delta \theta<\lambda_{bd,\theta} \quad \text{and}\quad \Delta r<\lambda_{bd,r}\\
-\infty, &\quad \text{otherwise}\\
\end{aligned}
\right.,
\end{split}
\end{equation}
where $\lambda_{bd,\theta}$ and $\lambda_{bd,r}$ are the potential parameters to be learned. They respectively describe the tolerance of the inter-frame offset and angle difference of the hypotheses.

Similarly, $\Phi_{ln}(\mathbf{h}_{ln,i-1},\mathbf{\hat{H}}_{ln,i},\mathbf{x}_{i})$ is defined as:
\begin{equation}
\begin{split}
\Phi_{ln,i}& = \phi(\mathbf{\hat{h}}_{ln1,i},\mathbf{x}_{ln,i})\\
&+ \phi(\mathbf{\hat{h}}_{ln2,i},\mathbf{x}_{ln,i}) + \varphi_{ln}(\mathbf{h}_{ln,i-1},\mathbf{\hat{h}}_{ln2,i})\\
&+\phi(\mathbf{\hat{h}}_{ln3,i},\mathbf{x}_{grad,i}) + \varphi_{ln}(\mathbf{h}_{ln,i-1},\mathbf{\hat{h}}_{ln3,i}),
\end{split}
\end{equation}

\subsubsection{Mode selection potential}
The mode selection potential $\Psi_{bd}(\mathbf{\hat{H}}_{bd,i},\mathbf{h}_{bd,i},\mathbf{x}_{i})$ seeks to select the best candidate bd hypothesis. A decision tree is used to guide the selection. Since the voting weights of candidates can indicate the hypotheses confidence, the decision tree takes such input to predict the best candidate. Let $\phi(\mathbf{\hat{h}}_{bd1,i},\mathbf{x}_{bd,i})$ be denoted as $\phi_{bd1}$ for short, the decision tree diagram for border is shown in Fig. \ref{Model_Decision_Tree}:
\begin{figure}[t,b,h]
  \centering
  \includegraphics[height = 3.4cm]{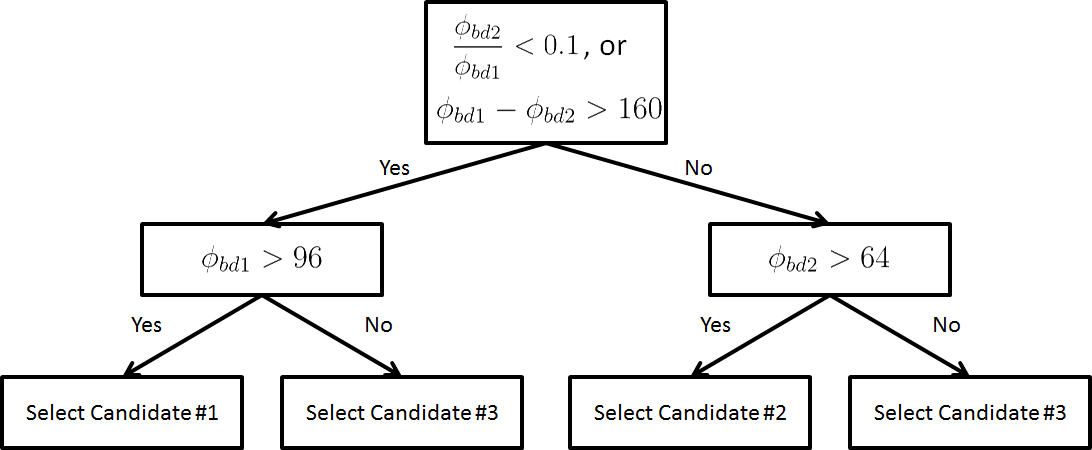}
\caption{Decision tree for border candidate hypothesis selection. The decision thresholds are selected based on empirical search for optimum performance.}\label{Model_Decision_Tree}
\end{figure}

Let $\mathbf{\hat{h}}_{bd,i}^{*}$ denote the candidate selected by the decision tree, and $\mathbf{\hat{H}}_{bd,i}^{C}$ the candidates not selected by the decision tree. The mode selection potential is defined as:
\begin{equation}
\Psi_{bd,i}=\left\{
\begin{aligned}
&0, &\text{if} \quad \mathbf{h}_{bd,i}=\mathbf{\hat{h}}_{bd,i}^{*}\\
&-\lambda_{mode}, &\text{if} \quad \mathbf{h}_{bd,i} \in \mathbf{\hat{H}}_{bd,i}^{C}\\
&-\infty, &\text{otherwise}\\
\end{aligned}
\right.,
\end{equation}
where $\lambda_{mode}$ is a nonnegative penalty parameter to be learned. It controls how sensitive the model is to the violation of decision tree output. The mode selection potential basically forces the output to be one of the candidate hypotheses, but allows discrepancy with the decision tree prediction with a penalty. The graphical model of the mode selection potential is shown in Fig. \ref{Model_Decompose} (b).

The decision tree for lane marking is similarly defined. The condition for the root decision node is chosen as: $\phi_{ln1}-\phi_{ln2}>50$ and the conditions for the two child decision nodes are respectively $\phi_{ln1}>16$ and $\phi_{ln2}>10$.

\subsubsection{Coupled structure potential}
The coupled structure potential $\Omega(\mathbf{h}_{bd,i},\mathbf{h}_{ln,i})$ further regularizes the results by exploiting the intra-frame structure. With this potential, border and lane marking are no longer independent but coupled and it can significantly improve the results under certain cases. The graphical representation of the coupled structure potential is illustrated in Fig. \ref{Model_Decompose} (c).

The potential mainly captures two properties of structural restrictions between border and lane marking:\\
\textbf{Parallelism:} The border and lane marking hypotheses are approximately parallel to each other. The closer they are, the stronger such property holds. Most important of all, they can not intersect.\\
\textbf{Distance:} A border often keeps certain distance from the lane marking. They can not be too close to each other.

Let $\Delta\theta_{i} \triangleq |\theta_{bd,i}-\theta_{ln,i}|$ and $\Delta r_{i} \triangleq r_{bd,i}-r_{ln,i}$. The coupled structure potential is defined as:
\begin{small}
\begin{equation}
\begin{split}
&\Omega(\mathbf{h}_{bd,i},\mathbf{h}_{ln,i})=\\
&\left\{
\begin{aligned}
0, \quad &\text{if} \quad \Delta r_{i}\geq D_{min}(r_{ln,i}), d_{1}\leq \Delta r_{i} <d_{2}, \Delta\theta_{i} \leq \lambda_{str1}\\
0, \quad &\text{if} \quad \Delta r_{i}\geq D_{min}(r_{ln,i}), d_{2}\leq \Delta r_{i} <d_{3}, \Delta\theta_{i} \leq \lambda_{str2}\\
0, \quad &\text{if} \quad \Delta r_{i} \geq d_{3}\\
-\infty, \quad &\text{otherwise}\\
\end{aligned}
\right.,
\end{split}
\end{equation}
\end{small}
where $\lambda_{str1}$ and $\lambda_{str2}$ are the potential parameters learned from training data to control the level of parallelism. $d_{1}$, $d_{2}$ and $d_{3}$ are empirically set to 10, 17 and 35 respectively. $D_{min}(r_{ln,i})$ is a piecewise-linear function defined as:
\begin{equation}
D_{min}(r_{ln,i}) = \max(\min(a r_{ln,i}+b,27),10),
\end{equation}
where $a$ and $b$ are parameters that can also be learned through a linear regression from training data.

\subsection{Inference}
The online updating nature (real-time requirement) of this problem limits our scope of observations to the past frames. Suppose at time $t$ the set of available observations is denoted as $\mathbf{X}_{1:t}$. The inference problem is to:
\begin{equation}
\mathbf{H}_{1:t}^{*}=\mathop{\arg\max}_{\mathbf{H}_{1:t}}P(\mathbf{H}_{1:t}|\mathbf{X}_{1:t}).
\end{equation}

However, conducting the above whole inference each time given a new frame is computationally infeasible. A relaxation is to initialize with the inferred state variable configuration of the previous $t-1$ frames and infer the current state variables, updating in an incremental way. At $t=1$, the inference is to optimize the following problem:
\begin{equation}
\begin{split}
&\mathbf{H}_{1}^{*} = \mathop{\arg\max}_{\mathbf{H}_{1}} \log P(\mathbf{H}_{1}|\mathbf{X}_{1})\\
 &=\mathop{\arg\max}_{\mathbf{h}_{bd,1},\mathbf{h}_{ln,1}}\phi(\mathbf{h}_{bd,1},\mathbf{x}_{bd,1}) + \phi(\mathbf{h}_{ln,1},\mathbf{x}_{ln,1})+\Omega_{1}
\end{split}
\end{equation}
This is to search the maximum Hough voting for border and lane marking based on Type 1 voting points, subject to the constraint from the coupled structure potential.

At $t>1$, the log probability can be represented as:
\begin{equation}
\begin{split}
P(\mathbf{H}_{1:t}&|\mathbf{X}_{1:t}) \propto \log P(\mathbf{H}_{1:t-1}|\mathbf{X}_{1:t-1})\\ &\exp(\Phi_{bd,t} + \Phi_{ln,t}+ \Psi_{bd,t} + \Psi_{ln,t} + \Omega_{t}).
\end{split}
\end{equation}
Since we reuse the previously inferred results, the inference problem becomes:
\begin{equation}
\begin{split}
\mathbf{H}_{t}^{*} &= \mathop{\arg\max}_{\mathbf{H}_{t}} \log P(\mathbf{H}_{1:t}|\mathbf{X}_{1:t})\\
 &=\mathop{\arg\max}_{\mathbf{H}_{t}} \Phi_{bd,t} + \Phi_{ln,t}+ \Psi_{bd,t} + \Psi_{ln,t} + \Omega_{t},
\end{split}
\end{equation}
where $\mathbf{H}_{t}\triangleq \{\mathbf{\hat{H}}_{bd,t},\mathbf{\hat{H}}_{ln,t},\mathbf{h}_{bd,t},\mathbf{h}_{ln,t}\}$.
The $-\infty$ penalty in the potential functions essentially serve as hard constraints on the search space of Hough voting hypotheses. To perform inference one needs to follow these constraints and conduct the following 4-step optimization:

\noindent{
\textbf{Step-1}: Optimize with $\mathbf{\hat{H}}_{bd,t}$ and $\mathbf{\hat{H}}_{ln,t}$.
}
\begin{equation}
\begin{split}
(\mathbf{\hat{H}}_{bd,t}^{*},\mathbf{\hat{H}}_{ln,t}^{*}) =\mathop{\arg\max}_{\mathbf{\hat{H}}_{bd,t},\mathbf{\hat{H}}_{ln,t}} \Phi_{bd,t} + \Phi_{ln,t}
\end{split}
\end{equation}
Given $\mathbf{h}_{bd,t-1}$ from the previous frame, generate the candidate hypothesis $\mathbf{\hat{h}}_{bd1,t}$ using unconstrained Hough voting and $\mathbf{\hat{h}}_{bd2,t}$ and $\mathbf{\hat{h}}_{bd3,t}$ using constrained Hough voting. Similar for lane marking candidate hypotheses.

\noindent{
\textbf{Step-2}: Optimize with $\mathbf{h}_{bd,t},\mathbf{h}_{ln,t}$.
}
\begin{equation}
\begin{split}
(\mathbf{h}_{bd,t}^{*},\mathbf{h}_{ln,t}^{*}) =\mathop{\arg\max}_{\mathbf{h}_{bd,t},\mathbf{h}_{ln,t}} \Psi_{bd,t} + \Psi_{ln,t}
\end{split}
\end{equation}
This step selects the candidate indicated by the decision tree for both border and lane marking.

\noindent{
\textbf{Step-3}: Back perturbation with $\mathbf{\hat{H}}_{bd,t}$. (Optional)
}

Check whether $\Omega(\mathbf{h}_{bd,t},\mathbf{h}_{ln,t})$ equals to $-\infty$. If it is then one needs to go back adjust $\mathbf{\hat{H}}_{bd,t}$ such that the three candidates also follow the structure restriction with $\mathbf{h}^{*}_{ln,t}$:
\begin{equation}
\begin{split}
(\mathbf{\hat{H}}_{bd,t}'^{*}) &= \mathop{\arg\max}_{\mathbf{\hat{H}}_{bd,t}} \Phi_{bd,t}+\Omega(\mathbf{\hat{h}}_{bd1,t},\mathbf{h}_{ln,t}^{*})\\
&+ \Omega(\mathbf{\hat{h}}_{bd2,t},\mathbf{h}_{ln,t}^{*}) + \Omega(\mathbf{\hat{h}}_{bd3,t},\mathbf{h}_{ln,t}^{*})
\end{split}
\end{equation}

\noindent{
\textbf{Step-4}: Optimize with $\mathbf{h}_{bd,t}$. (Optional)
}

Given the adjusted border hypotheses $\mathbf{\hat{H}}_{bd,t}$, again perform mode selection with the decision tree.

\subsection{Learning}
Much of the potential parameters in our model can be automatically learned from training data.
We use Gaussian models to fit the differences of $r$ and $\theta$ from ground truth hypotheses in consecutive frames. This gives a statistical estimate of how quick the hypotheses can change. Thus  $\lambda_{bd,\theta}$, $\lambda_{bd,r}$, $\lambda_{bd,\theta}$ and $\lambda_{bd,r}$ are learned as twice of the model standard deviation. The estimation of $\lambda_{str1}$ and $\lambda_{str2}$ is conducted similarly. $\lambda_{mode}$ is learned to be larger than the largest vote weight loss caused by back perturbation, such that one would not risk violating the decision tree to generate a candidate that does not follow the coupled structure restriction but with a larger voting weight.

\section{Implementation}
\subsection{Voting point extraction}
\begin{figure}[t,b,h]
  \centering
  \subfigure{
    \includegraphics[height = 1.25cm]{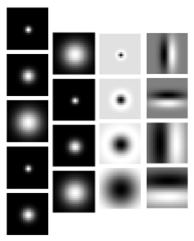}}
  \subfigure{
    \includegraphics[height = 1.25cm]{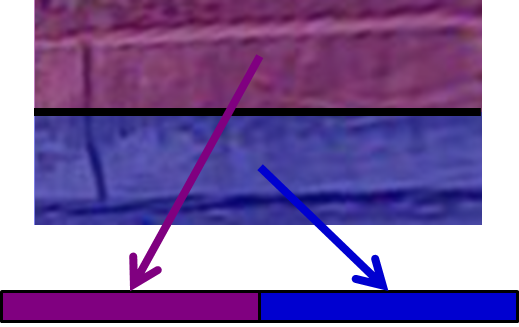}}
  \subfigure{
    \includegraphics[height = 1.25cm]{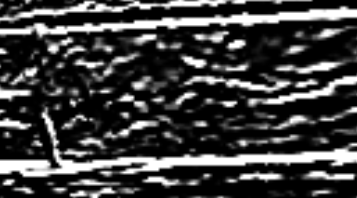}}
  \subfigure{
    \includegraphics[height = 1.25cm]{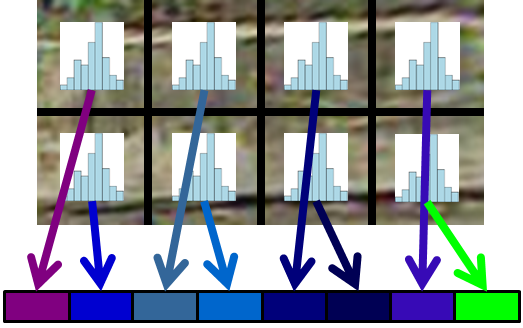}}
\caption{Illustration of feature extraction.}\label{Fig6}
\end{figure}

We first describe how to obtain the Type 1 voting points. We extract the filter bank responses and histogram of oriented gradient (HOG) to perform scanning window detection. The filter bank we used is the same as in \cite{12}, while the adopted HOG descriptor follows the work of \cite{47}. We divide each image patch into upper and lower two cells, and concatenate their mean filter bank responses. In addition, we also divide each image patch into 8 cells for HOG, where the normalized gradient histogram of all the cells are concatenated. The final feature for each detector window is the concatenation of the above filter bank responses and HOG features. Fig. \ref{Fig6} illustrates our feature extraction method.

It is worth mentioning that the computation of filter bank and HOG features for scanning windows can be conducted in an extremely efficient way using integral image. Thus the proposed feature extraction method has the potential to work real time.

We train two classifiers, one for border detection and the other for lane marker detection. In detail, we perform Fisher discriminant analysis on both the border training set and the lane marker training set, where the features of each training set are extracted as described previously. Then, we train two Radial Basis Function (RBF) Kernel SVMs on the dimensionality reduced training sets.

\subsection{Highway entrance and lane state detection}\label{Additional_Feature}
\begin{figure}[t,b,h]
  \centering
  \subfigure[]{
    \label{Fig6a}
    \includegraphics[height = 2cm]{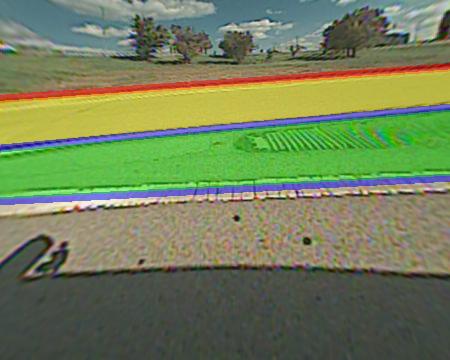}}
  \subfigure[]{
    \label{Fig6b}
    \includegraphics[height = 2cm]{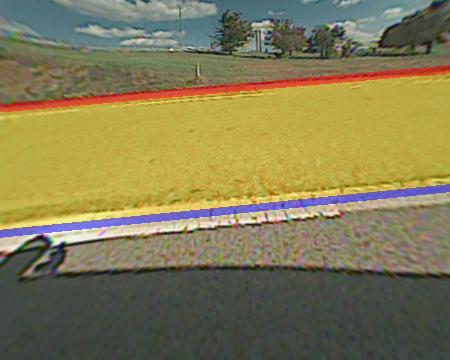}}
  \subfigure[]{
    \label{Fig6c}
    \includegraphics[height = 2cm]{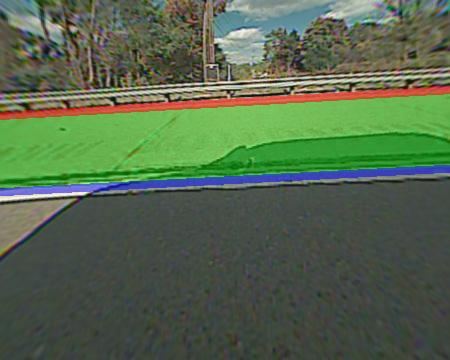}}
\caption{Examples of entrance detection and lane state tracking.}\label{AddFeat}
\end{figure}

In addition to the basic border and shoulder detection, we include highway entrance detection and lane state tracking which allow us to jointly estimate the position of merging/neighboring lane and tracking the lane state of the vehicle (e.g., whether it is the right-most lane). Figure \ref{AddFeat} shows some example results returned by the algorithm, where yellow regions in (a) and (b) indicate non-shoulder merging/neighboring lanes. The true shoulders on the other hand are detected as green regions in (c).

\section{Dataset}
\begin{figure}[t,b,h]
  \centering
  \subfigure[]{
    \includegraphics[height=2cm]{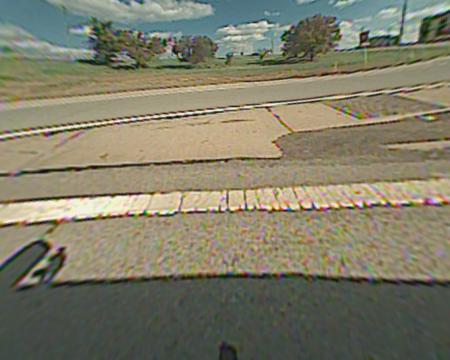}}
  \subfigure[]{
    \includegraphics[height=2cm]{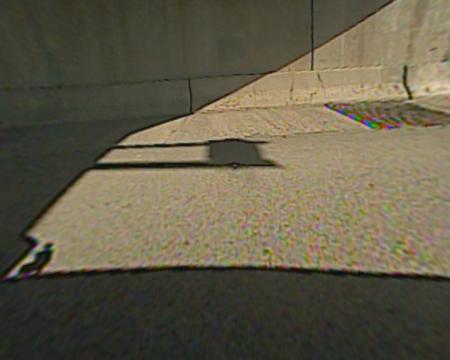}}
  \subfigure[]{
    \includegraphics[height=2cm]{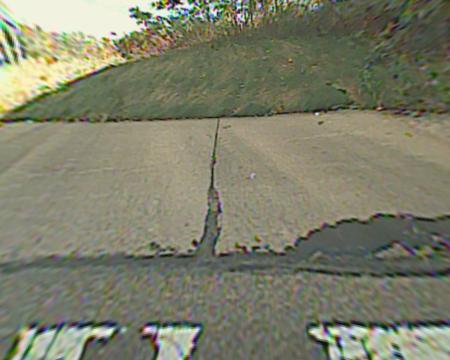}}
  \subfigure[]{
    \includegraphics[height=2cm]{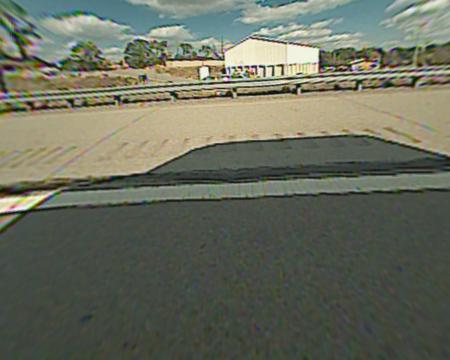}}
  \subfigure[]{
    \includegraphics[height=2cm]{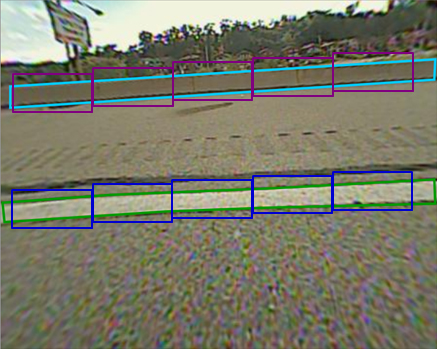}}
  \subfigure[]{
    \includegraphics[height=2cm]{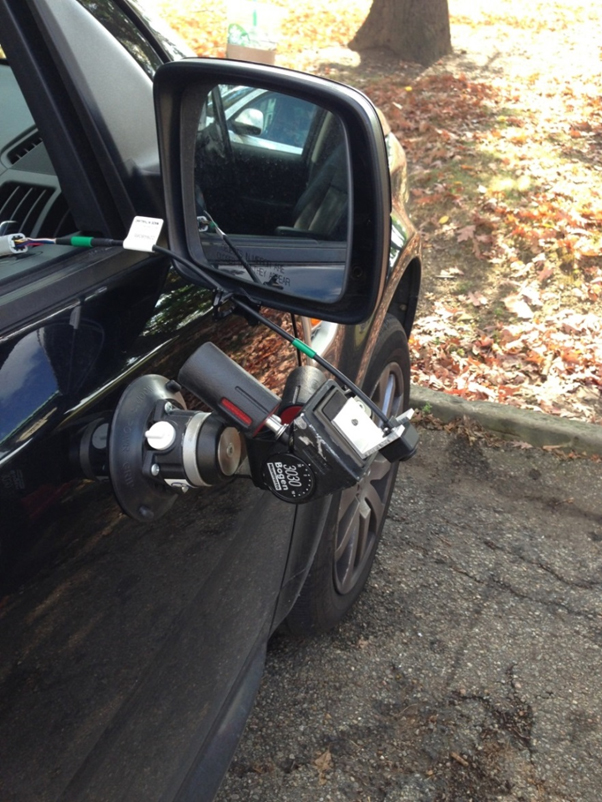}}
\caption{Examples data collection and labelling. (a)-(d) are example images from the collected dataset. (e) illustrates labeling and training patch alignment. (f) shows our system for data collection.}\label{Fig2}
\end{figure}

We collected 4200 highway road shoulder images. Among them, 1592 images are used for training where the images come from the frames of multiple video segments. The remaining 2608 images are used for testing and all the images form a complete video. The dataset contains many challenging and complicated scenarios. Example images from the dataset are shown in Fig. \ref{Fig2} (a)-(d). The number of training images containing concrete barriers, guard rails and soft shoulders are 839, 300 and 453, respectively.

\begin{figure}[t,b,h]
  \centering
    \subfigure{
    \includegraphics[height=0.9cm]{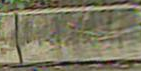}}
    \subfigure{
    \includegraphics[height=0.9cm]{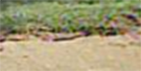}}
    \subfigure{
    \includegraphics[height=0.9cm]{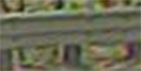}}
    \subfigure{
    \includegraphics[height=0.9cm]{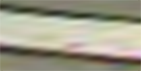}}
    \subfigure{
    \includegraphics[height=0.9cm]{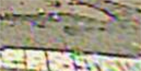}}
    \subfigure{
    \includegraphics[height=0.9cm]{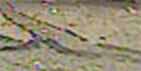}}
    \subfigure{
    \includegraphics[height=0.9cm]{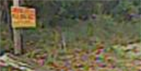}}
    \subfigure{
    \includegraphics[height=0.9cm]{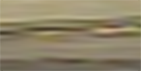}}
\caption{Examples of the training patches. Top row: positive patch examples. From left to right are respectively: concrete barrier, soft shoulder, guard rail and lane marker. Bottom row: mined negative samples}\label{Fig4}
\end{figure}

We use the MIT LabelMe open annotation tool to label borders and lane markings. We label the borders into ``concrete barriers'', ``guard rails'' and ``soft shoulders''. A set of well-aligned border image patches can be extracted from each annotated border and lane marker region (See Fig. \ref{Fig2} (e)). These patches form the positive training samples for the scanning window detectors, while negative samples are randomly mined from the background of the training images. The ratio between the number of positive samples and number of negative samples is set as $1-3$. Fig. \ref{Fig4} illustrates some examples of the training patches.

For test sequence, we label shoulder regions where the upper edges are the ground truth of highway borders. In the experimental section, we will use such ground truth to generate a set of benchmarks for quantitative comparison.

\section{Experimental results}
We conducted an experiment on the 2608 frames of the test video using structured Hough voting. The test sequence contains a variety of challenging situations, including complicated scenarios such as entrances and exits, as well as interfering visual manifestations such as strong shadows, dynamic appearances, drastic illumination change, weak border/lane marking and fake border/lane marking patterns.

To illustrate the performance of the proposed method, we compare our method with 3 baseline methods: 1. Independent Hough voting in each frame using the fired detector voting points, 2. Hough voting using the triggered detector voting points constrained by previous frame and 3. Adding gradient tracking to Baseline 2. We also compare to results obtained by the Kalman filter which is a standard method widely used in lane marking tracking.

\subsection{Adding coupled structure restrictions}
\begin{figure}[t,b,h]
  \centering
    \subfigure{
    \includegraphics[height=1.5cm]{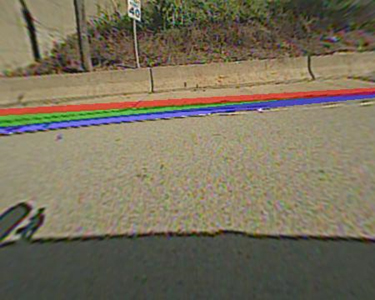}}
    \subfigure{
    \includegraphics[height=1.5cm]{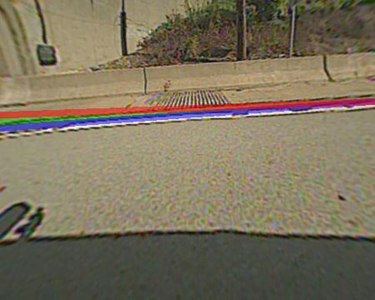}}
    \subfigure{
    \includegraphics[height=1.5cm]{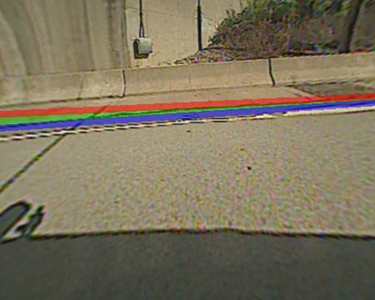}}
    \subfigure{
    \includegraphics[height=1.5cm]{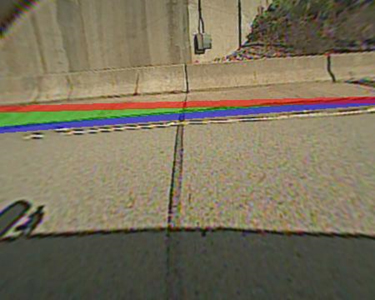}}
    \subfigure{
    \includegraphics[height=1.5cm]{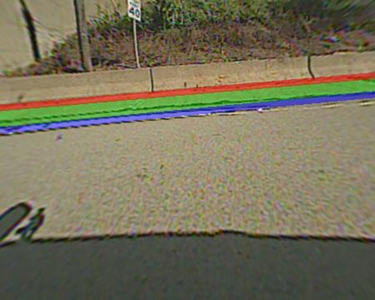}}
    \subfigure{
    \includegraphics[height=1.5cm]{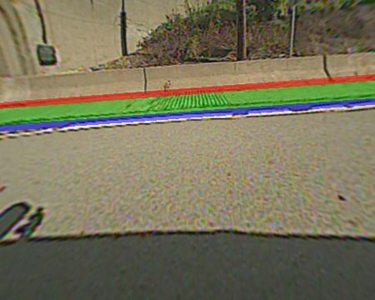}}
    \subfigure{
    \includegraphics[height=1.5cm]{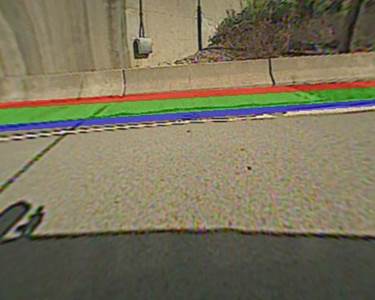}}
    \subfigure{
    \includegraphics[height=1.5cm]{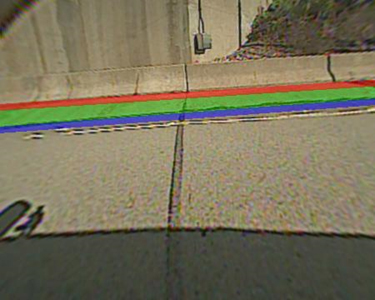}}
\caption{Examples of results successfully corrected by the coupled structure restriction.}\label{Fig10}
\end{figure}

We first illustrate examples that show the difference brought by the coupled structure restrictions. One can see that such restrictions have successfully corrected results where the detection of border failed. Some typical examples are illustrated in Fig. \ref{Fig10}. The first row corresponds to the method without coupled structure restriction, while the second row corresponds to the method with such restriction.

\subsection{Quantitative evaluation}
We conducted a series of evaluations based on the annotated test sequence ground truth and benchmarks\footnote{\textbf{Bd\_Pxl}: Average vertical pixel distortion between the detected border and hand annotated border ground truth.\\
\textbf{Ld\_Pxl}: Defined similarly to \textbf{Bd\_Pxl} for lane markings.\\
\textbf{Bd\_Ang}: Angle distortion between the detected border and the Hough configuration fit to the border ground truth.\\
\textbf{Ln\_Ang}: Defined similarly to \textbf{Bd\_Ang} for lane markings.\\
\textbf{Bd\_Pen}: Pixel distortion penalized by angle distortion:\\
$Bd\_Pen=\max(1, Bd\_Angle)*Bd\_Pxl$.\\
\textbf{Ln\_Pen}: Defined similarly to \textbf{Bd\_Pen} for lane markings.\\
\textbf{Accept\_Ratio}: Percentage of good frames defined by thresholding both \textbf{Bd\_Pen} and \textbf{Ln\_Pen}. A frame is ``good'' if both are within the threshold.\\
\textbf{Overlap\_Score}: $Score =\frac{Detection\_Region \cap Ground\_Truth}{Detection\_Region\cup Ground\_Truth}$.}. The quantitative results of the baselines and the proposed method (Proposed2 denotes with coupled structure restriction, while Proposed1 denotes without the restriction.) are listed in the table in Table \ref{Table1}. Again, one can see the performance of the proposed method is the best one.

\subsection{Qualitative evaluation}
\begin{figure*}[t,b,h]
  \centering
    \includegraphics[height=1.3cm]{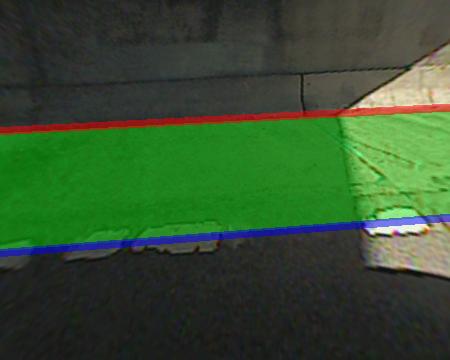}
    \includegraphics[height=1.3cm]{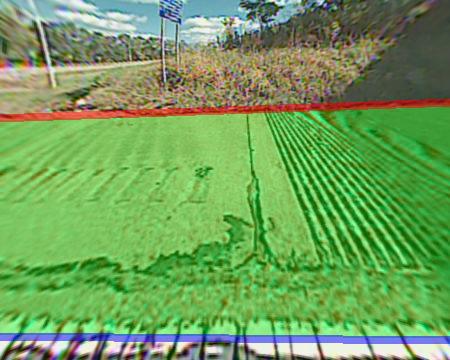}
    \includegraphics[height=1.3cm]{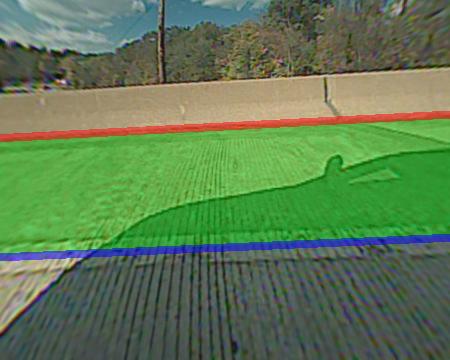}
    \includegraphics[height=1.3cm]{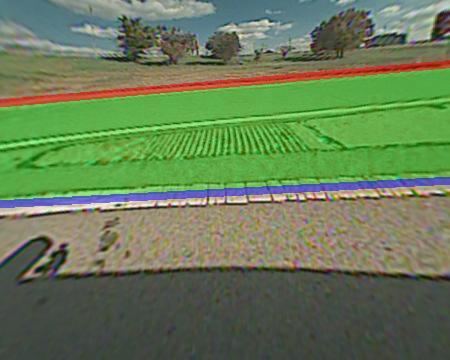}
    \includegraphics[height=1.3cm]{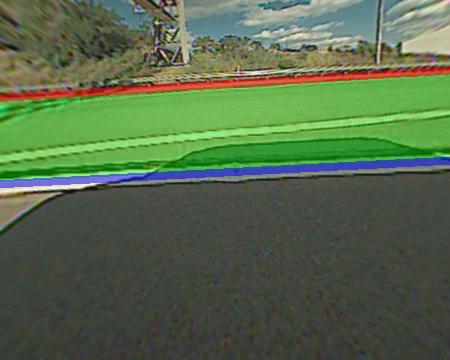}
\quad
    \includegraphics[height=1.3cm]{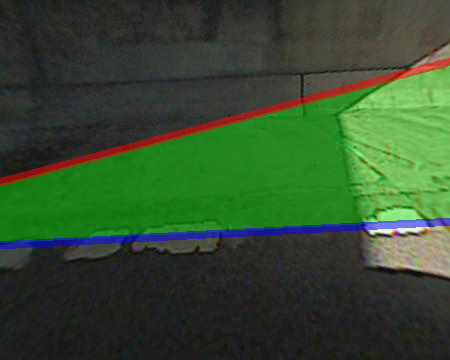}
    \includegraphics[height=1.3cm]{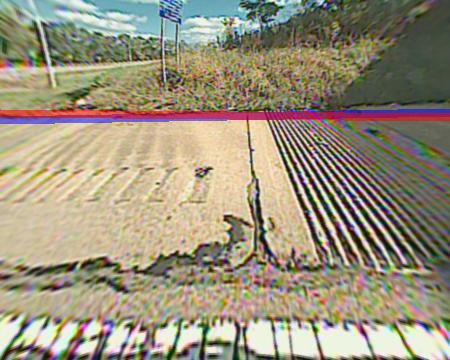}
    \includegraphics[height=1.3cm]{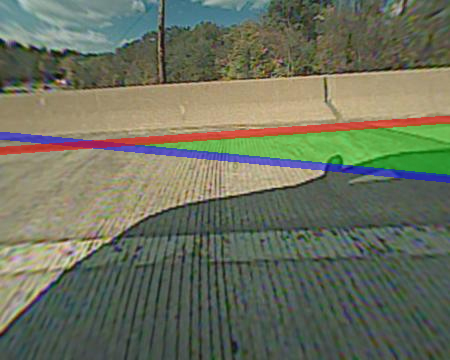}
    \includegraphics[height=1.3cm]{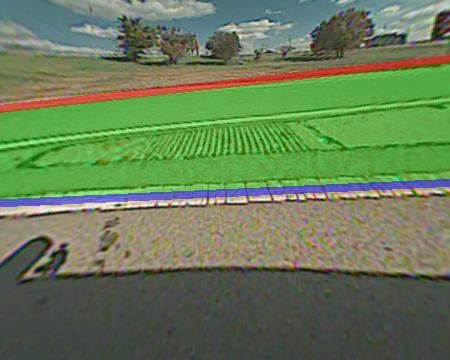}
    \includegraphics[height=1.3cm]{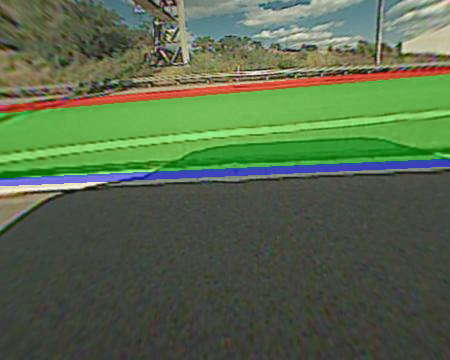}\\
\quad\\
    \includegraphics[height=1.3cm]{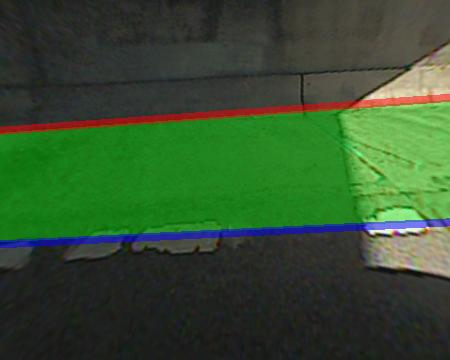}
    \includegraphics[height=1.3cm]{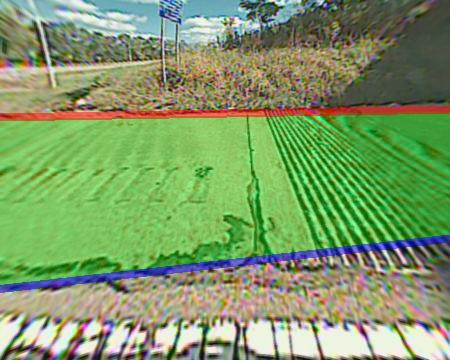}
    \includegraphics[height=1.3cm]{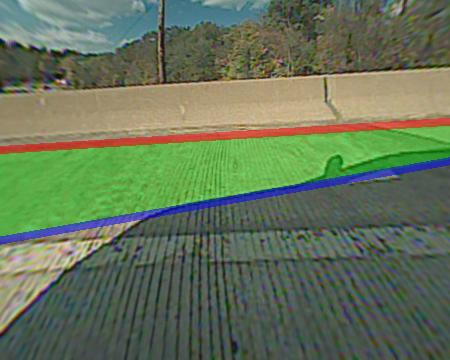}
    \includegraphics[height=1.3cm]{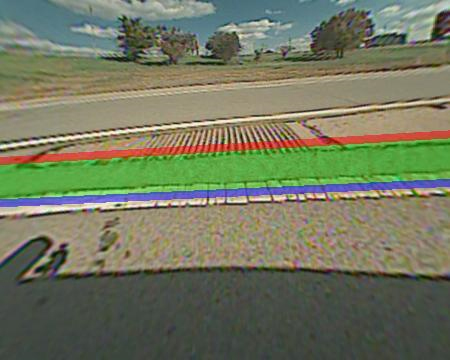}
    \includegraphics[height=1.3cm]{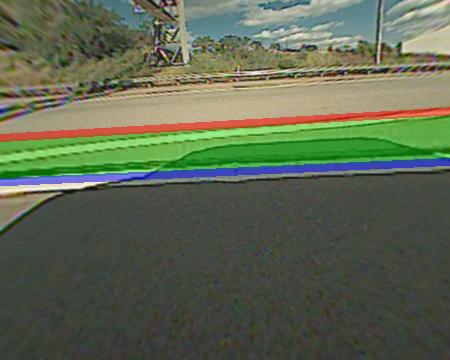}
\quad
    \includegraphics[height=1.3cm]{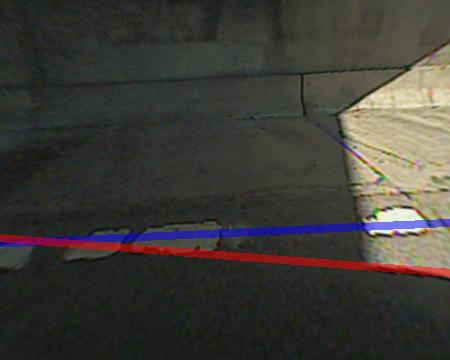}
    \includegraphics[height=1.3cm]{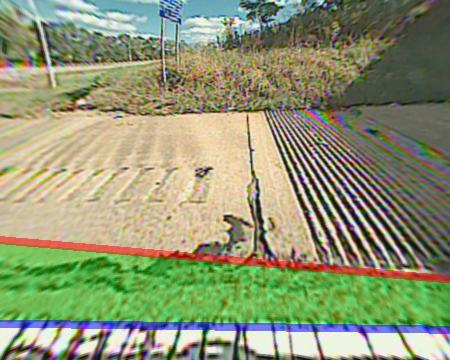}
    \includegraphics[height=1.3cm]{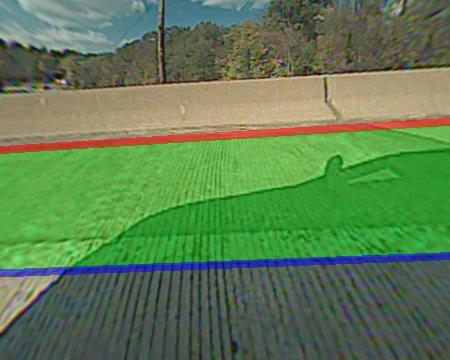}
    \includegraphics[height=1.3cm]{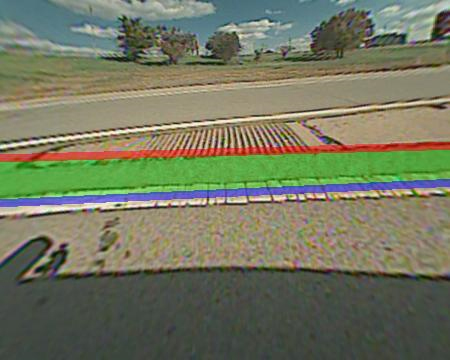}
    \includegraphics[height=1.3cm]{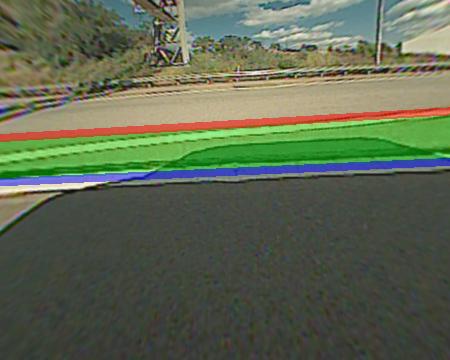}\\
\quad\\
    \includegraphics[height=1.3cm]{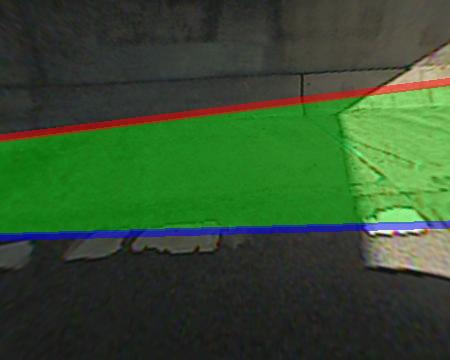}
    \includegraphics[height=1.3cm]{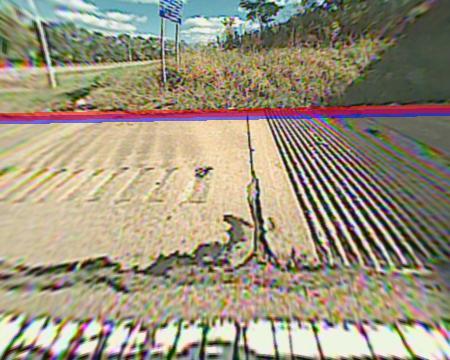}
    \includegraphics[height=1.3cm]{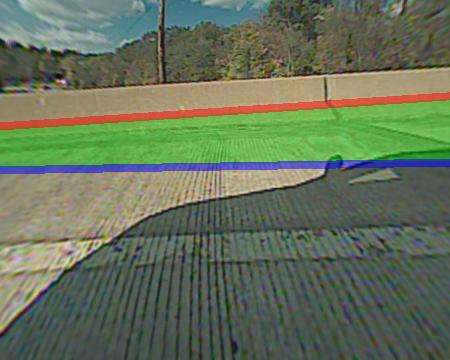}
    \includegraphics[height=1.3cm]{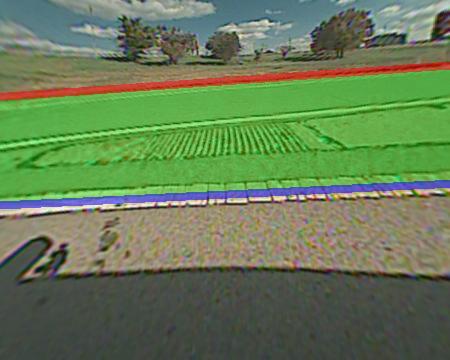}
    \includegraphics[height=1.3cm]{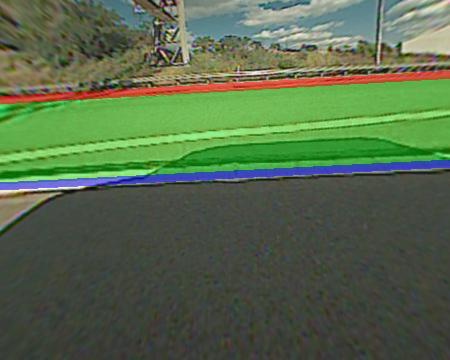}
\quad
    \includegraphics[height=1.3cm]{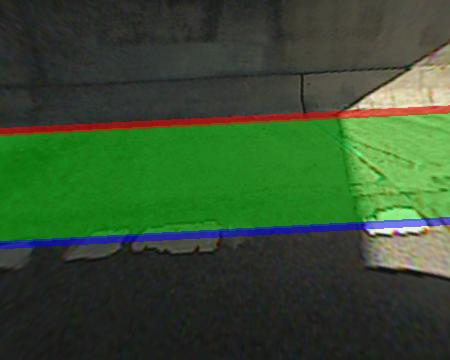}
    \includegraphics[height=1.3cm]{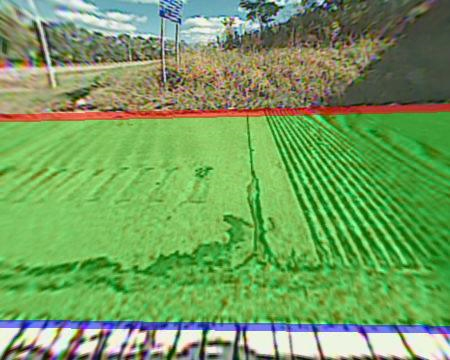}
    \includegraphics[height=1.3cm]{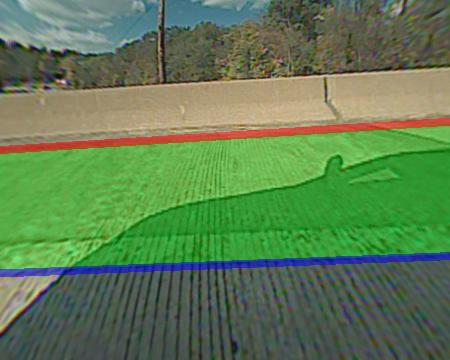}
    \includegraphics[height=1.3cm]{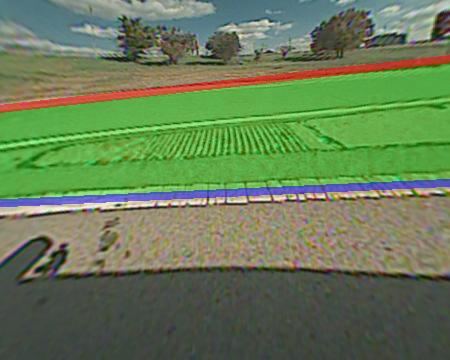}
    \includegraphics[height=1.3cm]{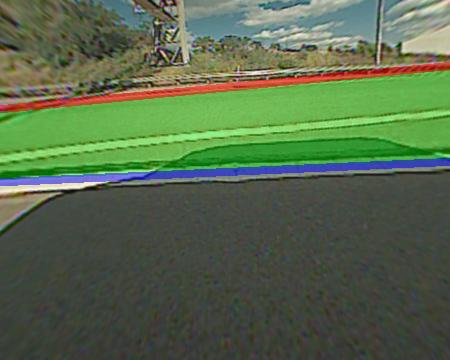}
\caption{Qualitative evaluation examples. The 5 images in each set of results respectively correspond to frame \#116, \#135, \#1011, \#36 and \#509. Note that results of the proposed method have not incorporated the entrance detection and lane tracking feature.}\label{Fig7}
\end{figure*}

We also select some challenging frames where the baseline methods usually fail. The results obtained by the baseline methods and the proposed method are shown in Fig. \ref{Fig7}. The 6 sets of images (from left to right and top to bottom) correspond to ground truth, baseline1, baseline2, baseline3, Kalman filter and the proposed method.

One can see that our model performs more robustly than the baseline methods in normal situations and is more responsive to drastic border changes in case of highway entrance due to the model flexibility achieved by the scheme of generating multiple hypothesis candidates.

\begin {table*}[t b h]
\caption {Quantitative segmentation evaluation}\label{Table1}
\begin{center}
\begin{tabular}{c|cccccc}
\hline
\hline
& \textbf{Baseline1} & \textbf{Baseline2} & \textbf{Baseline3} & \textbf{Kalman} & \textbf{Proposed1} & \textbf{Proposed2}\\
\hline
\textbf{Bd\_Pxl} & $3.4005$ & $3.8393$ & $16.2222$ & $3.6132$ & $2.9522$ &$\mathbf{2.7788}$\\
\textbf{Ld\_Pxl} & $8.1448$ & $8.8026$ & $7.4522$ & $8.9439$ & $\mathbf{3.6836}$ &$\mathbf{3.6836}$\\
\textbf{Bd\_Ang} & $0.6217$ & $0.5492$ & $1.0941$ & $\mathbf{0.4579}$ & $0.5201$ &$0.5103$\\
\textbf{Ln\_Ang} & $0.5891$ & $0.9711$ & $0.6729$ & $\mathbf{0.4851}$ & $0.5183$ &$0.5183$\\
\textbf{Bd\_Pen} & $6.0781$ & $5.4333$ & $74.4888$ & $4.4077$ & $3.7507$ &$\mathbf{3.4023}$\\
\textbf{Ln\_Pen} & $19.0167$ & $55.4144$ & $23.7144$ & $15.4963$ & $\mathbf{7.5538}$ &$\mathbf{7.5538}$\\
\textbf{Accept\_Ratio} & $0.8282$ & $0.8052$ & $0.7366$ & $0.8240$ & $0.8658$ &$\mathbf{0.8765}$\\
\textbf{Overlap\_Score} & $0.8984$ & $0.8905$ & $0.8125$ & $0.8915$ & $0.9302$ &$\mathbf{0.9335}$\\
\hline
\hline
\end{tabular}
\end{center}
\end {table*}

\subsection{Failure cases}
\begin{figure}[t,b,h]
  \centering
    \subfigure{
    \includegraphics[height=1.5cm]{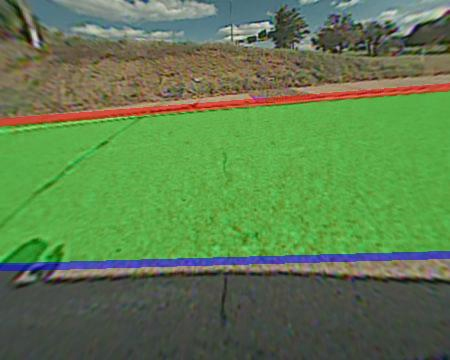}}
    \subfigure{
    \includegraphics[height=1.5cm]{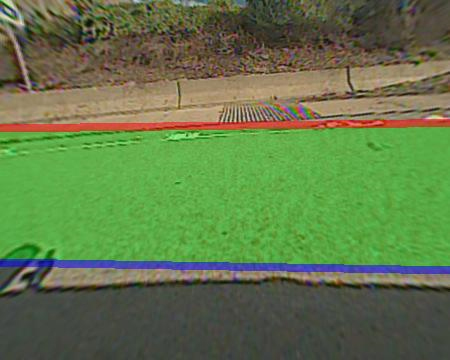}}
    \subfigure{
    \includegraphics[height=1.5cm]{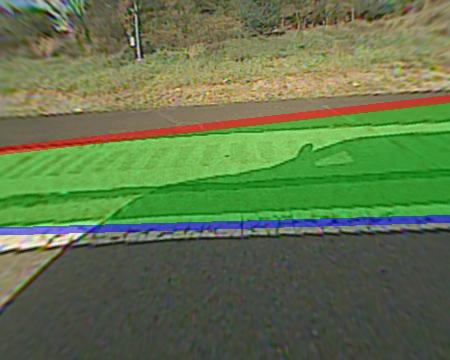}}
    \subfigure{
    \includegraphics[height=1.5cm]{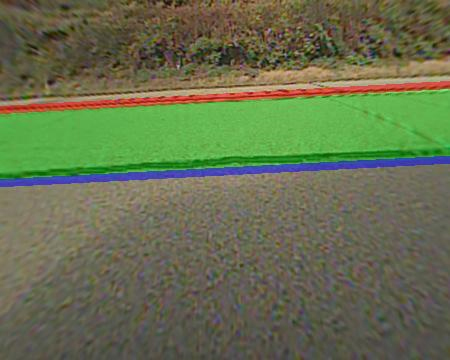}}
    \subfigure{
    \includegraphics[height=1.5cm]{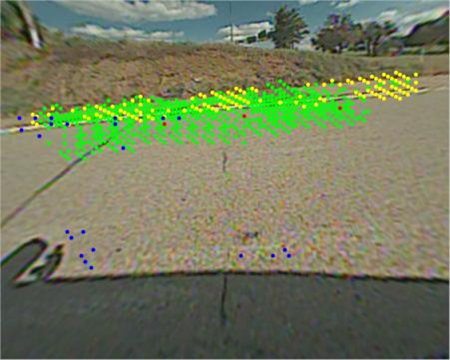}}
    \subfigure{
    \includegraphics[height=1.5cm]{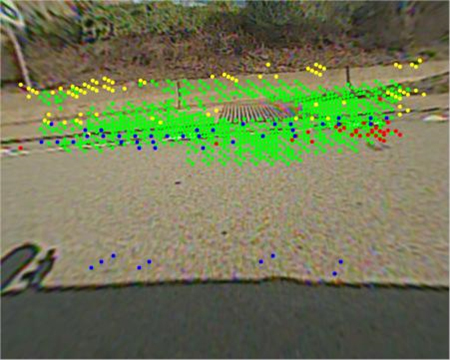}}
    \subfigure{
    \includegraphics[height=1.5cm]{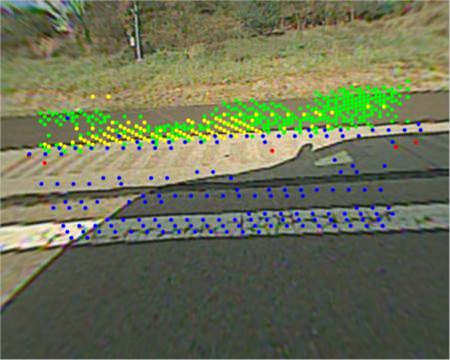}}
    \subfigure{
    \includegraphics[height=1.5cm]{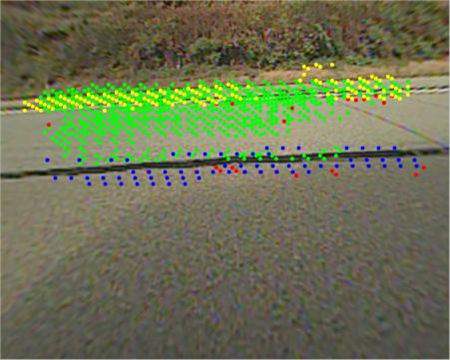}}
\caption{Some failure cases}\label{Fig9}
\end{figure}

We finally show some failure cases in Fig. \ref{Fig9}. The failure cases are mostly caused by false positive voting points and there is little a model could do given such input. This shows that the next step to the improve the method is to enhance the quality of voting point input.

\section{Conclusions}
In this paper, we have proposed a novel model called Structured Hough voting, and reported its application on vision-based highway border and shoulder detection. Experimental results have validated the good performance of the proposed model and its superiority over some popular models such as Kalman filter.

Our proposed method is also computationally efficient. First, the feature extraction unit can be implemented with integral image operations where the extraction of both mean filter bank responses and HOG in every scanning window is extremely fast. Second, our proposed inference method performs incremental (online) update, which also requires very few computation. The algorithm complexity of Hough voting is $\mathcal{O}(MN)$ where $M$ is the number of possible hypotheses and $N$ the number of voters. Voting with image gradients takes the most computation among the three candidate hypotheses generation methods. But it is still highly efficient since both hypotheses and voting points are significantly truncated by previous frame. In general, the method is able to run real-time without any GPU acceleration.


\small
\begin{thebibliography}{46}
\bibitem{2}
Z. Yu, W. Zhang and B.V.K. V. Kumar, ``Robust Rear-View Ground Surface Detection with Hidden State Conditional Random Field and Confidence Propagation,'' {\it ICIP} 2014.

\bibitem{6}
J. M. Alvarez, M. Salzmann and N. Barnes, ``Data Driven Road Detection,'' {\it WACV} 2014.

\bibitem{12}
J. Shotton, ``TextonBoost for image understanding, Multi-class object recognition and segmentation by jointly modeling texture, layout, and context,'' {\it IJCV}, 2007.

\bibitem{13}
M. Wilson et al., ``Poppet: A Robust Road Boundary Detection and Tracking Algorithm,'' {\it BMVC}, 1999.

\bibitem{14}
P. Charbonnier et al., ``Road Boundaries Detection Using Color Saturation,'' {\it Euro. Sig. Proc. Conf.}, 1998.

\bibitem{15}
S. Graovac and A. Goma, ``Detection of Road Image Borders based on Texture Classification,'' {\it Int. J. Adv. Robotic Sys.},2012.

\bibitem{16}
H. Kong, J.Y. Audibert and J. Ponce, ``General Road Detection From a Single Image,'' {\it IEEE Trans. IP}, 2010.

\bibitem{17}
J. Han, D. Kim, M. Lee and M. Sunwoo, ``Road Boundary Detection and Tracking for Structured and Unstructured Roads Using A 2D Lidar Sensor,'' {\it Int. J. Auto. Tech.}, 2014.

\bibitem{18}
A. Seibert, M. Hahnel, A. Tewes and R. Rojas, ``Camera based Detection and Classification of Soft Shoulders, Curbs and Guardrails,'' {\it IEEE IV}, 2013.

\bibitem{20}
H.Kong, J.Audibert and J. Ponce, ``Vanishing point detection for road detection,'' {\it CVPR}, 2009.

\bibitem{22}
Z. Sun, G. Bebis and R. Miller, ``On-Road Vehicle Detection,'' {\it IEEE Trans. PAMI}, 2006.

\bibitem{23}
R. Gopalan, T. Hong, M. Shneier, R. Chellappa, ``A Learning Approach Towards Detection and Tracking of Lane Markings,'' {\it IEEE Trans. ITS}, 2012.

\bibitem{25}
W. Choi, S. Savarese, ``Multiple Target Tracking in World Coordinate with Single, Minimally Calibrated Camera,'' {\it ECCV}, 2010.

\bibitem{30}
M. Bertozzi, A. Broggi, A. Fascioli and S. Nichele, ``Stereo visionbased vehicle detection,'' {\it IEEE IV}, 2000.

\bibitem{33}
E. Guizzo, ``How Google's Self-Driving Car Works,'' {\it IEEE Spectrum}, Feb. 26, 2013.

\bibitem{38}
D. Munoz, J. A. Bagnell, M. Hebert, ``Co-inference for Multi-modal Scene Analysis,'' {\it ECCV}, 2012.

\bibitem{41}
D. Munoz, J. A. Bagnell, M. Hebert, ``Stacked Hierarchical Labeling,'' {\it ECCV}, 2010.

\bibitem{44}
H. Cho et al., ``Vision-based 3D Bicycle Tracking using Deformable Part Model and Interacting Multiple Model Filter,'' {\it IEEE ICRA}, 2011.

\bibitem{47}
O. Ludwig, D. Delgado, V. Goncalves and U. Nunes, ``Trainable Classifier-Fusion Schemes: An Application To Pedestrian Detection,'' {\it IEEE ITSC}, 2009.

\bibitem{48}
W. Tao, X. He and N. Barnes, ``Learning Structured Hough Voting for Joint Object Detection and Occlusion Reasoning,'' {\it CVPR}, 2013.

\end{thebibliography}

\end{document}